\documentclass[sigconf]{acmart}

\usepackage{booktabs}
\usepackage{mathtools}
\usepackage{algorithm}
\usepackage{algpseudocode}
\usepackage{microtype}
\usepackage{xurl}
\usepackage{graphicx}
\usepackage[normalem]{ulem}
\usepackage{soul}

\newif\iftrackchanges
\trackchangesfalse  
\iftrackchanges
  \newcommand{\new}[1]{\textcolor{blue}{#1}}
  \newcommand{\removed}[1]{\textcolor{red}{\sout{#1}}}
\else
  \newcommand{\new}[1]{#1}
  \newcommand{\removed}[1]{}
\fi
\newtheorem{remark}{Remark} 
\graphicspath{{figures/}}
\AtBeginDocument{%
  }

\copyrightyear{2026}
\acmYear{2026}
\setcopyright{cc}
\setcctype{by}
\acmConference[GECCO '26]{Genetic and Evolutionary Computation Conference}{July 13--17, 2026}{San Jose, Costa Rica}
\acmBooktitle{Genetic and Evolutionary Computation Conference (GECCO '26), July 13--17, 2026, San Jose, Costa Rica}
\acmDOI{10.1145/3795095.3805174}
\acmISBN{979-8-4007-2487-9/2026/07}

\title{QUIVER: Cost-Aware Adaptive Preference Querying in Surrogate-Assisted Evolutionary Multi-Objective Optimization}

\author{Florian A. D. Burnat}
\affiliation{%
    \institution{University of Warwick}
    \department{Warwick Business School}
    \city{Coventry}
    \country{England, UK}}
\email{florian.burnat@warwick.ac.uk}

\begin{abstract}
Interactive multi-objective optimization systems face a budget allocation dilemma: one can spend resources on expensive objective evaluations or on eliciting decision-maker preferences that identify the relevant region of the Pareto set.
Moreover, preference elicitation itself spans modalities with different information content and cognitive burden, ranging from cheap, noisy pairwise preference statements (PS) to richer but costlier indifference adjustments (IA).

We study cost-aware optimization under an unknown \removed{scalarisation} \new{scalarization} and introduce QUIVER (Query-Informed Value Estimation for Regret), a surrogate-assisted evolutionary multi-objective optimizer that adaptively chooses between objective evaluations and heterogeneous preference queries.
At each step, QUIVER selects the next action by maximizing the expected decision-quality improvement per unit total cost.
Across DTLZ and WFG benchmarks under synthetic decision-maker models, QUIVER achieves the \emph{lowest} \new{final} utility regret on challenging WFG problems (\new{utility regret of} 2.14 on WFG4, 2.82 on WFG9: a 25\% improvement over baselines), outperforming all single-modality baselines.
We analyze how the optimal mix of PS and IA adapts to problem difficulty: on easy problems (DTLZ2), QUIVER selects 80\% PS queries; on hard problems (WFG9), it shifts to 35\% IA queries.
This adaptive modality selection demonstrates cost-aware preference learning in action.
\end{abstract}



\begin{document}
\maketitle

\section{Introduction}
\label{sec:intro}
Multi-objective optimization (MOO) seeks solutions that trade off multiple conflicting objectives.
In many real-world applications, from engineering design to portfolio selection, objective evaluations are expensive and involve physical experiments, costly simulations, or time-consuming computations.
This has motivated extensive research on surrogate-assisted and budgeted optimization methods that carefully allocate evaluation budget to efficiently approximate the Pareto front.

However, the Pareto front is typically infinite, and ultimately, a decision-maker (DM) must select a \emph{single} solution.
This selection depends on an \emph{unknown} scalar utility that captures the DM's preferences over trade-offs.
Without knowledge of these preferences, even a perfect approximation of the Pareto front may fail to identify the solution that the DM actually wants.
This creates a fundamental tension: \emph{objective evaluations} explore the Pareto set, whereas \emph{preference elicitation} identifies which region of that set matters.

\paragraph{The cost-aware allocation problem.}
Given a fixed total budget, how should an optimizer allocate resources between objective evaluation and preference elicitation?
Spending too much on evaluations may yield a well-approximated front in a region that the DM does not care about; spending too much on preference queries may identify the target region precisely but leave it poorly explored.
The optimal allocation depends on the relative costs and informativeness of each action, which vary across problem instances and over the course of optimization.

\paragraph{Heterogeneous preference modalities.}
Preference elicitation spans modalities with different information content and cognitive burden.
Pairwise preference statements (PS)---``Do you prefer $A$ or $B$?''---are cognitively cheap but yield only ordinal (1-bit) information.
Indifference adjustments (IA)---``How much would you adjust objective $k$ to make $A$ and $B$ equally preferred?''---are more demanding but reveal cardinal trade-off information, identifying the DM's marginal rate of substitution.
This heterogeneity raises a second allocation question: When should we use cheap, low-information queries versus expensive, high-information queries?

\paragraph{Novelty and approach.}
Most interactive EMO methods assume a fixed preference query protocol (e.g., periodic pairwise queries at fixed generations), whereas most surrogate-assisted EMO methods focus on Pareto front exploration and treat the DM as an \emph{a posteriori} selector.
In contrast, we treat \{\new{objective evaluation (}Eval\new{)}, PS, IA\} as a unified budgeted action space and introduce an adaptive controller that selects actions to maximize the expected decision-quality improvement \emph{per unit total cost}.
The value-of-information (VOI) perspective unifies objective exploration and preference learning under a single optimization criterion.

\paragraph{Optimizer-agnostic design.}
We designed the VOI-per-cost controller to be optimizer-agnostic.
In this study, we instantiate it with a surrogate-assisted NSGA-II and evaluate it using a protocol appropriate for evolutionary multi-objective optimization.
The controller requires only a minimal backend interface: (i) a mechanism to generate candidate designs $x$, (ii) storage for evaluated points, (iii) optionally a surrogate providing predictive distributions, and (iv) the ability to interleave preference queries with evaluations.
Any optimization backend satisfying these requirements, including Bayesian optimization, random search, or other population-based methods, can use the same VOI-per-cost action selection principle.

\paragraph{Contributions.}
This study makes four contributions.
\begin{enumerate}
  \item We \removed{formalise}\new{formalize} cost-aware interactive MOO with heterogeneous preference query modalities (PS and IA) under an unknown linear \removed{scalarisation}\new{scalarization}.
  \item We derive a practical VOI-based action selection rule using Monte Carlo estimation of expected entropy reduction.
  \item We propose QUIVER (\textbf{Qu}ery-\textbf{I}nformed \textbf{V}alue \textbf{E}stimation for \textbf{R}egret), an evolutionary algorithm with NSGA-II backbone that implements this adaptive allocation policy via a particle-filter preference posterior.
  \item We empirically evaluate on DTLZ and WFG benchmarks, demonstrating that QUIVER achieves lower utility regret per unit cost than fixed-schedule and single-modality baselines, with IA queries dominating when their information-per-cost ratio exceeds that of PS.
\end{enumerate}

\section{Related Work}
\label{sec:related}
\paragraph{Surrogate-assisted evolutionary multi-objective optimization.}
Expensive black-box multi-objective optimization has motivated extensive research on surrogate-assisted evolutionary algorithms.
ParEGO~\cite{Knowles2006-sx} scalarizes objectives via randomly sampled weight vectors and applies single-objective Bayesian optimization.
More recent approaches integrate surrogates directly into population-based EMO frameworks, including MOEA/D-EGO~\cite{Zhang2010-fz} and K-RVEA~\cite{Chugh2018-ko}.
Chugh et al.~\cite{Chugh2019-ey} categorized surrogate management strategies for expensive MOO.
These methods focus on efficiently approximating the Pareto front, but typically do not learn DM preferences online.

\paragraph{Interactive and preference-based EMO.}
Interactive evolutionary multi-objective optimization incorporates DM feedback to guide the search towards preferred regions.
Preference-based EMO learns implicit utilities from comparisons or rankings~\cite{Branke2015-pi,Brochu2007-bi}.
However, most approaches adopt fixed querying schedules and do not adaptively allocate budget between evaluations and preference elicitation.
Existing interactive EMO typically assume a fixed query budget or schedule, whereas we study joint allocation across evaluation and heterogeneous query types under a single cost budget.

\paragraph{Preference elicitation modalities and cognitive cost.}
Pairwise preference statements (PS) yield ordinal information and are commonly modelled using Bradley--Terry or probit likelihoods~\cite{Bradley1952-qt}.
Indifference adjustments (IA) provide cardinal trade-off information and can identify marginal rates of substitution in multi-attribute utility models~\cite{Keeney1993-ct}.
Recent empirical work by Haidinger et al.~\cite{Haidinger2025-wp} quantifies this trade-off: among 83 roofing professionals, IAs require only $\sim$15\% more time and cognitive effort than PSs, yet provide substantially richer information about cardinal preferences.
These differing cost--information trade-offs motivate adaptive strategies that choose query modalities based on the expected value per cost.

\paragraph{Cost-aware Bayesian optimization.}
Our cost-aware acquisition builds on the value-of-information (VOI) principles from decision analysis~\cite{Howard1966-ms,Raiffa1961-ao}.
In active learning, cost-sensitive methods select queries that maximize the information gain per unit cost~\cite{Settles2009-qx}.
A recent study on cost-aware BO~\cite{Lee2020-px} proposed acquisition functions that explicitly account for evaluation costs.
Our contribution extends cost-awareness to \emph{preference} queries, unifying objective evaluation and preference elicitation under a single, budgeted decision framework.

\paragraph{RLHF and preference optimization.}
Reinforcement learning from human feedback (RLHF)~\cite{Ouyang2022-rl} and direct preference optimization~\cite{Rafailov2023-dpo} rely almost exclusively on pairwise comparisons.
Our framework suggests that the strategic use of richer query types (analogous to IA) during critical training phases could reduce annotation budgets, which we discuss in Section~\ref{sec:discussion}.

\section{Problem Setup}
\label{sec:setup}
We consider the minimization of $m$ objectives $\mathbf{f}(x) \in \mathbb{R}^m$ over a feasible domain $\mathcal{X}$ (maximization problems are handled via sign-flip).
Objective evaluations are expensive and are counted against the cost budget.
A DM selects solutions according to an unknown utility $u(\mathbf{f}(x); w)$ parameterized by latent preferences $w$.

\subsection{Decision Objective: Decision Quality per Total Cost}
Let $x^\star$ denote the (unknown) utility-maximizing solution on the Pareto set, and let $\hat{x}$ be the algorithm's recommendation.
We evaluate decision quality via \emph{recommendation regret}:
\begin{equation}
\label{eq:regret}
\mathcal{R} = u(\mathbf{f}(x^\star_{\text{PF}}); w) - u(\mathbf{f}(\hat{x}); w).
\end{equation}
In the benchmarks, $x^\star_{\text{PF}}$ denotes the utility-optimal solution on the true Pareto front (computed analytically for DTLZ/WFG problems).
This metric is computed \emph{offline}: $x^\star_{\text{PF}}$ is the global optimum under $w^\star$, and is not limited to evaluated points.
We study optimization under a fixed total budget $B$ measured in cost units.

\paragraph{Note on regret metrics.}
Evolutionary algorithms can generate new solutions beyond any fixed candidate set; therefore, we measured regret against the true Pareto front optimal (computed analytically for benchmark problems).
Alternative backends (e.g., pool-based BO) that search over a finite candidate set may instead measure regret against the best point in that pool.
Because regret is defined relative to the analytic Pareto optimum, absolute regret values should not be compared across different optimization backbones.

\subsection{Actions and Costs}
At each step $t$, the optimizer chooses an action $a_t$ from:
\begin{itemize}
  \item \textbf{Eval}: evaluate objectives $\mathbf{f}(x)$ for a selected candidate $x$ (cost $c_{\text{eval}}$).
  \item \textbf{PS}: ask the DM to compare two outcomes $\mathbf{f}(x_i)$ and $\mathbf{f}(x_j)$ (cost $c_{\text{PS}}$).
  \item \textbf{IA}: ask the DM to adjust an attribute until indifference between two outcomes (cost $c_{\text{IA}}$).
\end{itemize}
The total spent cost must satisfy $\sum_t c(a_t) \le B$.
We define the \textbf{IA fraction} as the proportion of IA queries among all preference queries.
\begin{equation}
\text{IA fraction} = \frac{n_{\text{IA}}}{n_{\text{IA}} + n_{\text{PS}}},
\label{eq:ia_frac}
\end{equation}
where $n_{\text{IA}}$ and $n_{\text{PS}}$ are the counts of IA and PS queries, respectively.
This metric excludes objective evaluations and focuses on query modality selection.

\section{Preference Models}
\label{sec:prefmodels}
We describe the likelihood models for PS and IA and maintain a posterior $p(w\mid\mathcal{D}_{\text{pref}})$ over latent preferences.
For concreteness, we use a linear utility model $u(\mathbf{f};w)=w^\top \mathbf{f}$ with $w\in\Delta^{m-1}$, but the framework extends to other scalarizations.

\subsection{Pairwise Preference Statements (PS)}
A PS query returns $y\in\{0,1\}$ indicating whether outcome $A$ is preferred to $B$\new{, where $A = \mathbf{f}(x_i)$ and $B = \mathbf{f}(x_j)$ are objective vectors of two candidate solutions}.
We model noisy comparisons using a logistic likelihood:
\begin{equation}
\label{eq:pslik}
\Pr(y=1 \mid A,B,w) = \sigma\big(\beta\,[u(A;w) - u(B;w)]\big),
\end{equation}
where $\sigma(z)=(1+e^{-z})^{-1}$ and $\beta>0$ controls noise.

\subsection{Indifference Adjustments (IA)}
An IA query asks the DM to adjust one attribute (objective) until the two outcomes are equally preferred.
Let $e_k$ denote the $k$th unit vector.
Given outcomes $A$ and $B$, the DM reports an adjustment $\tilde{\delta}$ such that
\begin{equation}
\label{eq:iaindiff}
 u(A;w) \approx u(B + \tilde{\delta} e_k; w).
\end{equation}
Under linear utility, this implies
\begin{equation}
\label{eq:iaconstr}
 w^\top(A-B) \approx w_k\,\tilde{\delta}.
\end{equation}
We model the reported adjustment with additive noise as follows:
\begin{equation}
\label{eq:ialik}
\tilde{\delta} \mid (A,B,w,k) \sim \mathcal{N}\Big(\tfrac{w^\top(A-B)}{\max(w_k,\epsilon)},\,\sigma_{\text{IA}}^2\Big),
\end{equation}
where $\epsilon>0$ prevents numerical instability.
We treat IA as an abstract trade-off query that reveals an implied marginal rate of substitution along dimension $k$ rather than a literal objective edit.
This provides a cardinal signal about trade-offs and is typically more informative than PS, but incurs a higher cognitive cost.

\subsection{Posterior Inference over Preferences}
We maintain a posterior $p(w \mid \mathcal{D}_{\text{pref}})$ given the PS/IA observations using sequential Monte Carlo (SMC) with importance weighting.

\paragraph{Particle filter representation.}
We represent the posterior using $S$ weighted particles $\{(w^{(s)}, \omega^{(s)})\}_{s=1}^S$ where each $w^{(s)} \in \Delta^{m-1}$ is sampled from a Dirichlet prior $\text{Dir}(\alpha)$ at \removed{initialisation}\new{initialization}.
Upon observing a PS or IA response, we updated the particle weights via the corresponding likelihood as follows:
\begin{equation}
\omega^{(s)} \leftarrow \omega^{(s)} \cdot p(\text{observation} \mid w^{(s)}).
\end{equation}
When the effective sample size $\text{ESS} = 1/\sum_s (\omega^{(s)})^2$ \new{(a measure of how many particles carry meaningful weight; low ESS indicates weight degeneracy)} drops below a threshold (we use $S/2$), we resample particles with replacement proportional to weights and reset the weights to uniform.
This approach supports fast $O(S)$ updates and avoids the computational cost of the MCMC while providing a flexible nonparametric posterior approximation.

\section{QUIVER Method}
\label{sec:method}
QUIVER couples (i) an NSGA-II-based search layer that maintains and evolves a population of candidate solutions with (ii) a cost-aware controller that decides whether to spend budget on objective evaluations or preference queries.

\subsection{Search Layer: NSGA-II Backbone}
We use NSGA-II~\cite{Deb2002-ls} as the evolutionary backbone, maintaining a population $P_t$ of size $N$ and an archive $\mathcal{A}$ of evaluated solutions with their true objective values.

\paragraph{Variation operators.}
Offspring are generated using simulated binary crossover (SBX) with distribution index $\eta_c=20$ and polynomial mutation with distribution index $\eta_m=20$, following standard practice.

\paragraph{Environmental selection.}
After each evaluation, we updated the population using the NSGA-II selection mechanism: (1) combine parents and offspring, (2) perform fast non-dominated sorting to assign Pareto ranks, (3) compute crowding distance within each front, and (4) select the best $N$ individuals by rank (lower is better) with crowding distance as tiebreaker (higher is better).

\paragraph{Preference-biased selection.}
When recommending a final solution, we select the evaluated point that maximizes the posterior expected utility:
\begin{equation}
\hat{x} = \arg\max_{x \in \mathcal{A}} \mathbb{E}_{w \sim p(w|\mathcal{D}_{\text{pref}})}[u(\mathbf{f}(x); w)].
\end{equation}
This was approximated by averaging the particles.

\subsection{Cost-Aware Action Selection}
At each decision step, the controller selects an action $a$ from \{Eval, PS, IA\} by maximizing the expected improvement per unit cost:
\begin{equation}
\label{eq:policy}
 a^\star = \arg\max_{a} \frac{\mathbb{E}[\Delta Q \mid a,\mathcal{H}_t]}{c(a)},
\end{equation}
where $\mathcal{H}_t$ denotes the optimization history, and

\paragraph{Decision-quality proxy $Q$ and VOI definitions.}
We used posterior entropy over preferences as a proxy for decision quality.
Specifically, $Q = -H(p(w\mid\mathcal{D}_{\text{pref}}))$\new{, where $H(\cdot)$ denotes Shannon entropy (a scalar summary of posterior uncertainty: high entropy means the weight posterior is spread over many possible preference vectors, low entropy means it is concentrated)}, so that $\Delta Q$ corresponds to expected entropy reduction (information gain, IG).
This yields a unified VOI mapping as follows:
\begin{itemize}
  \item \textbf{Preference queries (PS and IA):} $\Delta Q = \text{IG}(\theta; \text{observation})$, the expected entropy reduction in the preference posterior.
  \item \textbf{Evaluation actions:} $\Delta Q = \text{VOC}_{\text{eval}}$, the expected improvement in posterior expected utility (a value-of-clairvoyance proxy).
\end{itemize}
This VOI decomposition is a general controller design choice, not specific to NSGA-II; any backend providing candidates and a preference posterior can use it.
Lower preference uncertainty concentrates the posterior mass over \removed{scalarisations}\new{scalarizations}, which improves the probability that the recommended solution maximizes DM utility.

\paragraph{Estimating expected information gain.}
For preference queries (PS and IA), we estimate $\mathbb{E}[\Delta Q]$ using a Monte Carlo simulation (Algorithm~\ref{alg:eig}).
We sampled possible query outcomes from the current posterior, computed the resulting posterior update, and measured the entropy reduction.

\paragraph{Binned mutual information for IA.}
A technical subtlety arises when comparing information gain across modalities: IA queries yield continuous responses, whereas PS queries yield binary responses.
The raw entropy-based MI for continuous outcomes can be artificially inflated.
We address this by discretizing IA outcomes into $B=15$ bins before computing the information gain:
\begin{equation}
\tilde{\delta}_{\text{binned}} = \text{bin\_center}(\text{digitize}(\tilde{\delta}, \text{edges})).
\end{equation}
This binned MI approach ensures a fair comparison between modalities while preserving the ordinal information content of the IA responses.
For evaluations, we use a decision-aligned value: let $U_{\text{best}} = \max_{x \in \mathcal{A}} \mathbb{E}_w[u(\mathbf{f}(x); w)]$ be the current best posterior expected utility among the evaluated points.
The evaluation value is $\text{VOC}_{\text{eval}} = \max(0, U_{\text{cand}} - U_{\text{best}})$, where $U_{\text{cand}}$ is the expected utility of the best candidate.
This makes the evaluation VOC comparable to the preference query VOC: both measure the expected improvement in decision quality.
\new{Although IG and VOC$_{\text{eval}}$ measure different quantities (entropy reduction vs.\ utility improvement), both are proxies for the same underlying goal: reducing recommendation regret. The cost-\removed{normalised}\new{normalized} ratio in Eq.~\eqref{eq:policy} compares their expected contributions to this shared objective, making cross-modality comparison meaningful.}

\begin{algorithm}[t]
\caption{QUIVER (high-level)}
\label{alg:quiver-ea}
\begin{algorithmic}[1]
\State \removed{Initialise}\new{Initialize} population $P$, empty evaluation archive $\mathcal{A}$, preference data $\mathcal{D}_{\text{pref}}\leftarrow\emptyset$
\State Spend initial budget on a small batch of \textbf{Eval} to seed $\mathcal{A}$
\While{spent cost $\le B$}
  \State \textit{// Backend: generate candidates (EA-specific)}
  \State Fit/update surrogate $\hat{\mathbf{f}}$ using archive $\mathcal{A}$ \new{\Comment{Optional; omit if no surrogate}}
  \State Generate offspring candidates $C$ from $P$ \new{via variation operators (optionally with surrogate pre-screening)}
  \State \textit{// Controller: VOI/cost action selection (optimizer-agnostic)}
  \State Select action $a\in\{\text{Eval},\text{PS},\text{IA}\}$ using Eq.~\eqref{eq:policy}
  \State \textit{// Backend: execute action and update models}
  \If{$a=\textbf{Eval}$}
    \State Choose $x\in C$ (e.g., surrogate-guided infill), evaluate $\mathbf{f}(x)$, update $\mathcal{A}$
  \ElsIf{$a=\textbf{PS}$}
    \State Choose pair $(x_i,x_j)$, query preference label $y$, update $\mathcal{D}_{\text{pref}}$
  \Else
    \State Choose IA query $(x_i,x_j,k)$, observe $\tilde{\delta}$, update $\mathcal{D}_{\text{pref}}$
  \EndIf
  \State Update preference posterior $p(w\mid\mathcal{D}_{\text{pref}})$
  \State Update population $P$ (e.g., NSGA-II selection biased to preferred region)
\EndWhile
\new{\Ensure Recommended solution $\hat{x}$}
\State \new{\Return} $\hat{x}$ \removed{maximising}\new{maximizing} posterior expected utility
\end{algorithmic}
\end{algorithm}

\begin{algorithm}[t]
\caption{Estimating Expected Information Gain}
\label{alg:eig}
\begin{algorithmic}[1]
\Require Query $(A, B)$ for PS, or $(A, B, k)$ for IA; particles $\{(w^{(s)}, \omega^{(s)})\}$; $M$ Monte Carlo samples
\new{\Ensure Estimated expected information gain EIG}
\State $H_0 \leftarrow$ entropy of current particle weights
\State $\text{EIG} \leftarrow 0$
\For{$i = 1$ to $M$}
  \State Sample $w^* \sim p(w|\mathcal{D}_{\text{pref}})$ \Comment{Sample from posterior}
  \State Simulate observation $o \sim p(o | w^*, \text{query})$ \Comment{PS: $y \in \{0,1\}$; IA: $\tilde{\delta} \in \mathbb{R}$}
  \State $\{\tilde{\omega}^{(s)}\} \leftarrow$ update weights given $o$ \Comment{Temporary copy}
  \State $H_{\text{after}} \leftarrow$ entropy of $\{\tilde{\omega}^{(s)}\}$
  \State $\text{EIG} \leftarrow \text{EIG} + (H_0 - H_{\text{after}}) / M$
\EndFor
\State \Return EIG
\end{algorithmic}
\end{algorithm}

\paragraph{Query pair selection.}
For preference queries, we select pairs $(A, B)$ from the evaluated archive using \emph{posterior disagreement}, an active learning approach.
For each candidate pair, we sampled $n$ weight vectors from the posterior and computed the fraction $p$ that preferred $A$ over $B$.
The disagreement score is $\min(p, 1-p)$, maximized at $p=0.5$ when the posterior is maximally uncertain of the preference.
We select the pair with the highest disagreement, ensuring that the queries are informative rather than trivial.
For IA queries, we additionally select the objective $k$ with the highest posterior variance $\text{Var}[w_k]$, targeting the most uncertain dimension of the weight vector.

\subsection{Theoretical Analysis}
\label{sec:theory}

We provide a theoretical grounding for the VOI-per-cost policy, establishing when each query modality is preferred.

\paragraph{Information content of query modalities.}
The informativeness of PS and IA queries differs fundamentally.
Under linear utility $u(f;w) = w^\top f$, a PS query comparing outcomes $A$ and $B$ yields Fisher information:
\begin{equation}
\label{eq:fisher_ps}
I_{\text{PS}}(w) = \frac{(w^\top(A - B))^2}{\sigma_{\text{PS}}^2} \cdot p(1-p),
\end{equation}
where $p = \sigma(w^\top(A-B)/\sigma_{\text{PS}})$ is the probability of preferring $A$.
This information vanishes when the DM is nearly indifferent ($p \approx 0.5$) or strongly opinionated ($p \approx 0$ or $1$).

In contrast, an IA query adjusting objective $k$ yields the Fisher information:
\begin{equation}
\label{eq:fisher_ia}
I_{\text{IA}}^{(k)}(w) = \frac{1}{\sigma_{\text{IA}}^2},
\end{equation}
independent of the current preference estimate.
This explains IA's advantage: its informativeness does not depend on query difficulty, whereas PS information content varies with the preference gap.

\paragraph{Modality switching threshold.}
The VOI-per-cost policy in Eq. ~\eqref{eq:policy} induces a natural threshold for the query modality selection.

\begin{remark}[Modality switching threshold] 
\label{prop:threshold}
Under the cost-\removed{normalised}\new{normalized} policy, QUIVER selects IA over PS if and only if
\begin{equation}
\frac{\text{IG}_{\text{IA}}}{\text{IG}_{\text{PS}}} > \frac{c_{\text{IA}}}{c_{\text{PS}}}.
\end{equation}
\new{This follows directly from Eq.~\eqref{eq:policy}: IA is selected when $\text{IG}_{\text{IA}}/c_{\text{IA}} > \text{IG}_{\text{PS}}/c_{\text{PS}}$.}
\end{remark}

This threshold has an intuitive interpretation: IA queries are worthwhile only when their relative information advantage exceeds their relative cost disadvantage.
For the empirically grounded cost ratio $c_{\text{IA}}/c_{\text{PS}} = 1.15$~\cite{Haidinger2025-wp}, QUIVER requires IA to provide at least 15\% more information than PS to justify selection.
Our cost sensitivity experiments (Section~\ref{sec:results}) confirm a crossover near $c_{\text{IA}}/c_{\text{PS}} \approx 2$, indicating that IA provides approximately twice the information of PS under our noise model.

\section{Experimental Setup}
\label{sec:experiments}

\paragraph{Benchmarks.}
We evaluate on standard DTLZ~\cite{Deb2005-di} and WFG~\cite{Huband2006-li} test suites:
\begin{itemize}
  \item \textbf{DTLZ2} ($m=3, 5$): Concave Pareto front, tests convergence and distribution.
  \item \textbf{WFG4} ($m=3$): Multi-modal, separable, concave front.
  \item \textbf{WFG9} ($m=3$): Non-separable, multi-modal, deceptive---a challenging test case.
\end{itemize}
The decision space dimension was $d = m + 9$ for DTLZ and $d = 2m + 20$ for WFG, following the standard practice.
\new{While our experiments focus on $m \in \{3, 5\}$ objectives, the QUIVER framework (preference posterior, VOI/cost controller, and action selection) is general and applies to any number of objectives $m \geq 2$.}

\paragraph{Cost model.}
We set costs as: $c_{\text{eval}} = 5.0$, $c_{\text{PS}} = 1.0$, and $c_{\text{IA}} = 1.15$ (15\% higher than PS).
The IA cost premium is grounded in empirical findings from \cite{Haidinger2025-wp}, who studied 83 roofing professionals and found that indifference adjustments require only $\sim$15\% more response time and cognitive effort than pairwise comparisons.
The total budget is $B = 500$ cost units, allowing approximately 100 evaluations or 435 PS queries.

\paragraph{Synthetic DMs.}
We sampled ground-truth preference weights $w^\star$ from $\text{Dir}(1, \ldots, 1)$ and simulated noisy responses:
\begin{itemize}
  \item PS: Bradley-Terry with noise parameter $\sigma_{\text{PS}} = 0.15$
  \item IA: Gaussian noise with $\sigma_{\text{IA}} = 0.18$ (20\% higher than PS)
\end{itemize}
Higher IA noise reflects the greater difficulty in specifying exact indifference adjustments.

\paragraph{Baselines.}
We compared against five baselines sharing the same NSGA-II backbone:
\begin{enumerate}
  \item \textbf{Eval-only}: Standard surrogate-assisted EMO without preference learning. The final recommendation uses uniform weights $w = (1/m, \ldots, 1/m)$.
  \item \textbf{PS-only}: All preference budget spent on pairwise statements with random pair selection.
  \item \textbf{IA-only}: All preference budget spent on indifference adjustments with random pair/objective selection.
  \item \textbf{Fixed-schedule}: Alternates Eval, PS, IA at fixed intervals.
  \item \textbf{Random}: Samples actions proportional to inverse cost (with cheaper actions more likely).
\end{enumerate}
The Eval-only baseline uses uniform weights for recommendation because it has no preference information; this is a fair baseline because any other fixed weighting would be arbitrary.

\paragraph{Algorithm parameters.}
Population size $N=20$, $S=2048$ particles for the preference posterior, $M=50$ Monte Carlo samples for EIG estimation.
All experiments were run for five independent seeds with different $w^\star$ samples.

\paragraph{Metrics.}
We report:
\begin{itemize}
  \item \textbf{Utility regret} $\mathcal{R}$ (Eq.~\ref{eq:regret}): the difference between the optimal and recommended utilities
  \item \textbf{Query mix}: fraction of budget spent on Eval, PS, and IA.
  \item \textbf{Final entropy}: posterior uncertainty at termination.
\end{itemize}

\section{Results}
\label{sec:results}

\subsection{Main Benchmark Results}

Tables~\ref{tab:main} and~\ref{tab:actions} summarize the final utility regret and query modality selection across all benchmarks.
Key findings:
\begin{itemize}
  \item \textbf{DTLZ2: All methods achieve near-zero regret.} On DTLZ2 ($m=3,5$), all methods, including Eval-only, achieve essentially zero regret ($\leq 0.02$), indicating that this benchmark is relatively easy with a sufficient budget.
  \item \textbf{WFG benchmarks reveal method differences.} Clear differences emerged in the more challenging WFG4 and WFG9 problems. QUIVER achieves the \emph{lowest} regret on both (2.14 and 2.82, respectively), outperforming all baselines, including PS-only and IA-only.
  \item \textbf{25\% improvement on WFG9.} On the hardest benchmark (WFG9), QUIVER achieves a regret of 2.82 vs. 3.77--3.90 for baselines, a 25\% reduction.
  \item \textbf{Adaptive modality selection.} QUIVER's VOI-per-cost policy adapts modality selection to problem difficulty: on easy problems (DTLZ2), it uses 80--87\% PS queries; on hard problems (WFG9), it shifts to 35\% IA queries to gather richer information.
\end{itemize}

\begin{table}[t]
\caption{Final utility regret (mean $\pm$ std, 5 seeds, $B$=500). Lower is better. Bold indicates best per row.}
\label{tab:main}
\centering
\scriptsize
\begin{tabular}{lcccc}
\toprule
Problem & Eval-only & PS-only & IA-only & QUIVER \\
\midrule
DTLZ2 (3) & $.02 \pm .04$ & $\mathbf{.00}$ & $\mathbf{.00}$ & $\mathbf{.00}$ \\
DTLZ2 (5) & $.00 \pm .01$ & $\mathbf{.00}$ & $\mathbf{.00}$ & $\mathbf{.00}$ \\
WFG4 (3) & $2.43 \pm .12$ & $2.24 \pm .13$ & $2.24 \pm .13$ & $\mathbf{2.14 \pm .15}$ \\
WFG9 (3) & $3.77 \pm .91$ & $3.90 \pm 1.5$ & $3.90 \pm 1.5$ & $\mathbf{2.82 \pm 1.4}$ \\
\bottomrule
\end{tabular}
\end{table}

\begin{table}[t]
\caption{Action counts for QUIVER (mean over 5 seeds, $B=500$). Shows the number of PS queries (n\_ps), IA queries (n\_ia), and IA fraction among preference queries. The policy adapts modality selection to problem difficulty.}
\label{tab:actions}
\centering
\small
\begin{tabular}{lcccc}
\toprule
Problem & n\_ps & n\_ia & IA fraction \\
\midrule
DTLZ2 ($m$=3) & 28.8 & 7.2 & 20\% \\
DTLZ2 ($m$=5) & 31.4 & 4.6 & 13\% \\
WFG4 ($m$=3) & 26.6 & 9.4 & 26\% \\
WFG9 ($m$=3) & 23.4 & 12.6 & 35\% \\
\bottomrule
\end{tabular}
\end{table}

\subsection{Query Mix Analysis}

\begin{figure}[t]
\centering
\Description{Bar chart showing utility regret for four policies across DTLZ2 and WFG benchmarks. Eval-only shows 23-34 percent regret while preference-learning methods show near-zero regret.}
\includegraphics[width=\columnwidth]{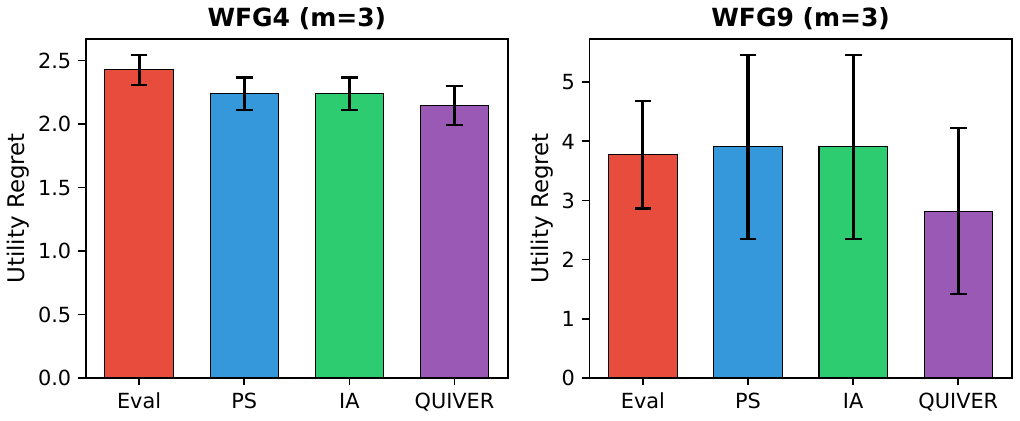}
\caption{Utility regret comparison across benchmarks. Preference-learning methods achieve near-zero regret while Eval-only has 23--34\% regret.}
\label{fig:regret}
\end{figure}

Figure~\ref{fig:regret} visualizes the regret comparison across all benchmarks.
On DTLZ2, all the methods achieved near-zero regret.
On WFG4 and WFG9, QUIVER achieved the lowest regret, demonstrating the value of adaptive action selection in challenging problems.

\begin{figure}[t]
\centering
\Description{Stacked bar chart showing QUIVER budget allocation across benchmarks. Evaluations consume approximately 92 percent of the cost budget, with the remaining budget split between PS and IA queries. IA fraction among preference queries ranges from 20 to 35 percent depending on problem difficulty.}
\includegraphics[width=0.85\columnwidth]{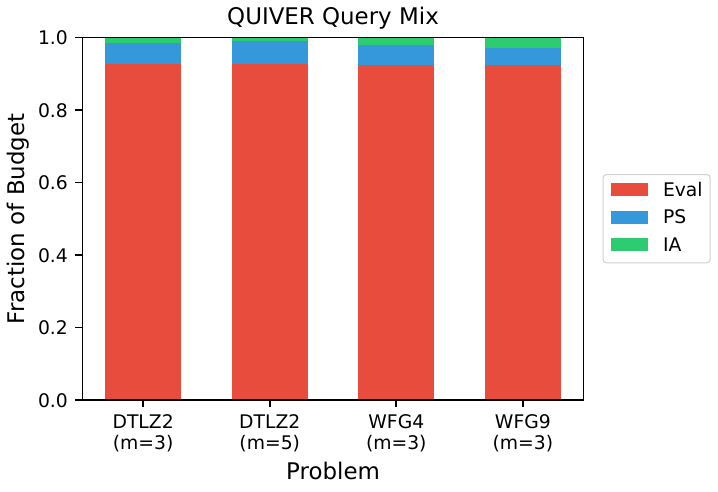}
\caption{Budget allocation for QUIVER across benchmarks. Evaluations dominate the cost budget ($\sim$92\%), with IA fraction among preference queries ranging from 20\% (easy DTLZ2) to 35\% (hard WFG9).}
\label{fig:querymix}
\end{figure}

Figure~\ref{fig:querymix} shows the budget allocation for QUIVER.
The policy adapts its modality selection based on the problem difficulty.
\begin{itemize}
  \item On DTLZ2 ($m$=3): 29 PS, 7 IA (20\% IA fraction---easy problem, cheap PS queries suffice).
  \item On WFG4: 27 PS, 9 IA (26\% IA fraction---moderate difficulty).
  \item On WFG9: 23 PS, 13 IA (35\% IA fraction---hard problem, richer IA queries preferred).
\end{itemize}
This demonstrates the core contribution: QUIVER \emph{automatically} shifts to more informative (but costlier) IA queries on harder problems, where the additional information is worth the cost.

\subsection{Cost Ratio Sensitivity}

\begin{figure}[t]
\centering
\Description{Line plot showing fraction of IA queries versus cost ratio. IA fraction decreases from 23 percent at cost ratio 1.0 to 10 percent at cost ratio 3.0.}
\includegraphics[width=0.85\columnwidth]{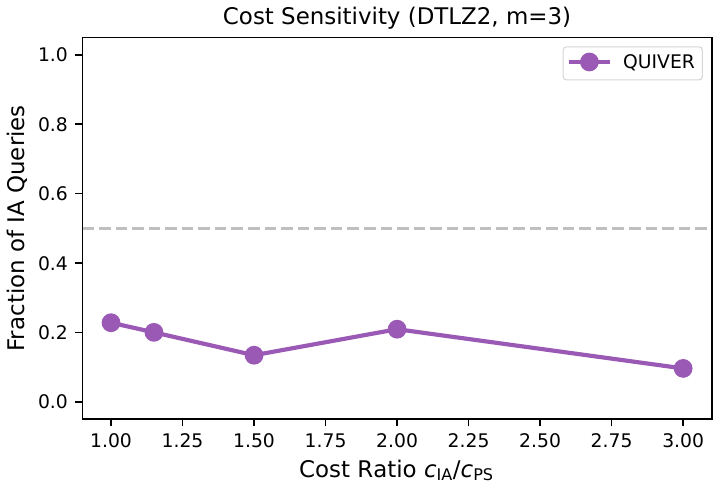}
\caption{Cost sensitivity analysis on DTLZ2 ($m=3$). QUIVER gradually reduces IA usage as the cost ratio increases, demonstrating cost-aware modality selection.}
\label{fig:cost}
\end{figure}

To test robustness, we varied the IA cost ratio $c_{\text{IA}}/c_{\text{PS}}$ from 1.0 to 3.0 on DTLZ2 ($m=3$).
Figure~\ref{fig:cost} shows how QUIVER's query selection adapts to the cost structure.
As IA becomes more expensive, the IA fraction decreases:
\begin{itemize}
  \item At $c_{\text{IA}}/c_{\text{PS}} = 1.0$: IA fraction is 23\% (IA provides good value).
  \item At $c_{\text{IA}}/c_{\text{PS}} = 1.5$: IA fraction drops to 13\%.
  \item At $c_{\text{IA}}/c_{\text{PS}} = 3.0$: IA fraction falls to 10\% (PS dominates).
\end{itemize}
This gradual decline demonstrates that the VOI/cost policy automatically adapts to the cost structure, progressively shifting towards cheaper PS queries as IA becomes relatively more expensive.

\paragraph{Setting-dependent behavior.}
The observed modality usage is protocol dependent (benchmark geometry, noise, budget, and candidate construction); therefore, we report our conclusions within this experimental setup.
In QUIVER's evolutionary setting on DTLZ2, PS queries dominated across all cost ratios (77--90\% of preference queries), with IA usage decreasing as its relative cost increased.
The key finding is that VOI-per-cost control adapts query modality usage as costs evolve, automatically selecting the most cost-efficient modality.

\subsection{Fatigue Regime: Dynamic IA Cost}
\label{sec:fatigue}

To demonstrate true adaptive behavior, we implemented a \emph{fatigue regime} where the IA cost increases with usage:
\begin{equation}
c_{\text{IA}}(t) = c_{\text{IA}}^0 \times (1 + \alpha \times n_{\text{IA}}(t)),
\end{equation}
where $\alpha$ is the fatigue coefficient, and $n_{\text{IA}}(t)$ is the cumulative IA query count.
This models cognitive fatigue, where complex queries become progressively more burdensome.

Table~\ref{tab:fatigue} shows the results for DTLZ2 ($m=3$) with varying fatigue coefficients.
Key findings:
\begin{itemize}
  \item \textbf{Adaptive switching demonstrated.} As fatigue $\alpha$ increases, the IA fraction decreases from 62\% to 55\%, showing that the policy responds to increasing costs.
  \item \textbf{Reduced total queries.} Higher fatigue reduces overall preference queries ($n_{\text{IA}} + n_{\text{PS}}$ drops from 159 to 70) as the algorithm shifts budget toward evaluations when preference queries become expensive.
  \item \textbf{Graceful degradation.} Regret increased gradually from 0\% to 14\% with higher fatigue, not catastrophically.
\end{itemize}

\begin{table}[t]
\caption{Fatigue experiment results on DTLZ2 ($m=3$, 5 seeds). As fatigue coefficient $\alpha$ increases, QUIVER adaptively reduces IA usage.}
\label{tab:fatigue}
\centering
\small
\begin{tabular}{lcccc}
\toprule
Fatigue $\alpha$ & Regret & IA Fraction & $n_{\text{IA}}$ & $n_{\text{PS}}$ \\
\midrule
0.00 & $0.00 \pm 0.00$ & 62\% & 99 & 60 \\
0.05 & $0.03 \pm 0.06$ & 58\% & 53 & 39 \\
0.10 & $0.13 \pm 0.21$ & 54\% & 47 & 42 \\
0.15 & $0.14 \pm 0.18$ & 55\% & 38 & 32 \\
\bottomrule
\end{tabular}
\end{table}

\section{Discussion}
\label{sec:discussion}

\paragraph{Problem difficulty matters.}
On DTLZ2, all methods achieved near-zero regret, suggesting that this benchmark is relatively easy when the budget is sufficient.
The real test comes on WFG4 and WFG9, where QUIVER achieves the \emph{lowest} regret among all methods (2.14 and 2.82, respectively).
Notably, on these harder problems, even PS-only and IA-only with 347--400 preference queries achieve higher regret than QUIVER, demonstrating the value of adaptive action selection.

\paragraph{Adaptive allocation outperforms fixed strategies.}
QUIVER adapts its action mix to the problem at hand: on DTLZ2, it uses 92 evals with mostly PS queries (20\% IA), whereas on WFG9, it shifts to 35\% IA queries to gather richer information.
This problem-dependent allocation explains why QUIVER achieves lower regret on harder problems: it recognizes when preference learning is more valuable and allocates accordingly.
Fixed strategies cannot make this adaptation, leading to suboptimal performance when budget allocation does not match the problem structure.

\new{\paragraph{Baselines as ablation study.}
Our comparisons are between QUIVER and its own internal variants (Eval-only, PS-only, IA-only, Fixed-schedule, Random), all sharing the same NSGA-II backbone.
This ablation design isolates the contribution of adaptive VOI/cost action selection from confounds introduced by different optimization backends.
Comparison with existing interactive EMO methods (e.g., NAUTILUS, NIMBUS, or preference-based MOEA/D variants) would require controlling for differences in search strategy, population management, and stopping criteria, which is beyond the scope of this study.
We view such cross-method comparison as important future work, noting that the VOI/cost controller is optimizer-agnostic and could be combined with these alternative backends.}

\paragraph{Limitations.}
Our experiments used synthetic DMs with linear utilities.
Real DMs may exhibit inconsistencies or non-linear preferences that our model does not capture.
\new{%
\paragraph{Linear utility assumption.}
The linear scalarization $u(\mathbf{f}; w) = w^\top \mathbf{f}$ cannot represent preferences over non-convex regions of the Pareto front, because linear utility maximization can only recover solutions on the convex hull of the Pareto set.
For problems with non-convex or disconnected Pareto fronts, this assumption may exclude the DM's true preferred solution from the reachable set.
Extending QUIVER to Chebyshev scalarization $u(\mathbf{f}; w) = \max_k w_k |f_k - z_k^*|$ or general non-linear utility models would require replacing the closed-form IA likelihood (Eq.~\ref{eq:ialik}) with a simulation-based likelihood, but the VOI/cost controller and particle-filter posterior remain applicable without modification.
We view this extension as a priority for future work.
}%
We address fatigue via a dynamic cost model (Section~\ref{sec:fatigue}), which shows that the framework can accommodate time-varying cognitive costs.
The particle filter posterior requires $O(S)$ storage and update cost, although this is negligible compared to objective evaluations in expensive optimization settings.
The VOI policy's reliance on entropy-based information gain may not capture all aspects of decision quality, and integrating surrogate predictions for candidates could improve action selection.

\paragraph{Human factors and cognitive modelling.}
Our cost model treats $c_{\text{PS}}$ and $c_{\text{IA}}$ as fixed constants, but real cognitive costs may vary with user fatigue, query difficulty, and individual differences in fatigue and difficulty.
\new{Moreover, fatigue may increase not only the cost but also the \emph{noise} of responses: a fatigued DM may provide less precise indifference adjustments ($\sigma_{\text{IA}}$ increases) and more erratic pairwise comparisons ($\sigma_{\text{PS}}$ increases), reducing the information content of each query.
Jointly modelling cost and noise degradation is an important direction for realistic deployment.}
The fatigue regime (Section~\ref{sec:fatigue}) demonstrates that the framework can accommodate dynamic costs.
Future work could model individual-specific cost functions learned online, enabling personalized query selection that accounts for heterogeneity across decision-makers.
Strategic or inconsistent users, who may misreport preferences or violate transitivity, present another challenge that robust preference learning techniques could address.

\new{\paragraph{Computational complexity.}
The per-step cost of QUIVER's action selection is dominated by the EIG estimation (Algorithm~\ref{alg:eig}): $O(M \cdot S)$ for $M$ Monte Carlo samples over $S$ particles, evaluated for each candidate action.
With $M=50$ and $S=2048$, this yields $\sim$$10^5$ lightweight operations per decision step---negligible compared to the wall-clock cost of an expensive objective evaluation.
The NSGA-II backbone adds $O(N^2 m)$ per generation for non-dominated sorting.
Overall, the computational overhead of adaptive action selection is small relative to the evaluation budget it manages.}

\paragraph{Connection to information-directed sampling.}
QUIVER's cost normalized acquisition can be viewed through the lens of information-directed sampling (IDS)~\cite{Russo2018-zc}, which balances information gain against instantaneous regret.
Our formulation extends this principle to heterogeneous query types: rather than trading off exploration vs.\ exploitation within a single query modality, we trade off across modalities with different information-cost profiles.
The modality-switching threshold (Remark~\ref{prop:threshold}) provides an interpretable rule that is analogous to the information ratio in the IDS.

\paragraph{Implications for RLHF efficiency.}
The principles underlying \new{our method}\removed{QUIVER} (adaptively selecting query modalities to maximize information per unit cost) have direct implications for reinforcement learning from human feedback (RLHF).
Current RLHF pipelines rely almost exclusively on pairwise comparisons; however, our results suggest that the strategic use of richer query types (analogous to IA) during critical training phases could reduce annotation budgets.
For example, when fine-tuning is nearly complete and the model needs to distinguish between high-quality responses, more informative (but costlier) queries may be justified in this case.
The cost-normalized VOI framework provides a principled way to make such decisions.

\paragraph{Extensions.}
The framework naturally extends to (i) additional query modalities (e.g., ranking queries, reference point specification), (ii) non-linear utility models via appropriate likelihood functions, (iii) cognitive budget constraints that limit total preference queries independent of cost, and (iv) improved Eval VOI using surrogate-predicted candidate utilities.

\section{Conclusion}
\label{sec:conclusion}
We introduced QUIVER, a cost-aware interactive evolutionary multi-objective optimizer that adaptively allocates budget between objective evaluations and heterogeneous preference queries (PS and IA).
The key innovation is a VOI/cost action selection rule that maximizes the expected information gain per unit cost, implemented via Monte Carlo estimation over a particle-filter preference posterior.

Our main finding is that \textbf{adaptive modality selection outperforms fixed strategies for challenging problems}.
On DTLZ2, all methods achieve near-zero regret.
However, on the harder WFG4 and WFG9 benchmarks, QUIVER achieves the \emph{lowest} regret (2.14 and 2.82, respectively---a 25\% improvement on WFG9), outperforming both Eval-only and fixed preference strategies.
The key advantage is QUIVER's adaptive modality selection: for easy problems, it uses cheap PS queries (80\% PS); for hard problems, it shifts to richer IA queries (35\% IA), where the additional information is worth the cost.

Across our baselines, the consistent finding is that VOI-per-cost control improves utility regret per unit cost and adapts query modality usage as costs and uncertainty evolve.
The framework demonstrates that treating \{Eval, PS, IA\} as a unified action space with different costs enables principled resource allocation in interactive multi-objective optimization.
This cost-normalized action selection principle is the key innovation, which is applicable regardless of the specific optimization backend.
Future work will investigate improved Eval VOI estimation using surrogate predictions, validate the findings with real human DMs\new{---whose fatigue patterns, inconsistencies, and learning effects may differ substantially from our synthetic models}---, and extend to non-linear utility models.

\bibliographystyle{ACM-Reference-Format}
\bibliography{gecco}

\end{document}